\documentclass[final]{cvpr}

\usepackage{times}
\usepackage{epsfig}
\usepackage{graphicx}
\usepackage{amsmath}
\usepackage{amssymb}

\usepackage[pagebackref=true,breaklinks=true,colorlinks,bookmarks=false]{hyperref}

\def\cvprPaperID{****} 
\def\confYear{CVPR 2021}

\begin{document}

\title{Wisdom \emph{for} the Crowd: Discoursive Power in Annotation Instructions for Computer Vision} 

\author{Milagros Miceli\\
Weizenbaum Institute\\
Germany\\
{\tt\small m.miceli@tu-berlin.de}
\and
Julian Posada\\
University of Toronto\\
Canada\\
{\tt\small julian.posada@mail.utoronto.ca}
}

\maketitle

\begin{abstract}
 Developers of computer vision algorithms outsource some of the labor involved in annotating training data through business process outsourcing companies and crowdsourcing platforms. Many data annotators are situated in the Global South and are considered independent contractors. This paper focuses on the experiences of Argentinian and Venezuelan annotation workers. Through qualitative methods, we explore the discourses encoded in the task instructions that these workers follow to annotate computer vision datasets. Our preliminary findings indicate that annotation instructions reflect worldviews imposed on workers and, through their labor, on datasets. Moreover, we observe that for-profit goals drive task instructions and that managers and algorithms make sure annotations are done according to requesters' commands. This configuration presents a form of commodified labor that perpetuates power asymmetries while reinforcing social inequalities and is compelled to reproduce them into datasets and, subsequently, in computer vision systems.
\end{abstract}

\section{Introduction}

Computer vision systems are built from training data previously collected, cleaned, and annotated by human workers. Companies and research institutions outsource many of these tasks through online labor platforms \cite{Posada2020a} and business process outsourcing (BPO) companies \cite{miceli_between_2020}. In these instances, outsourcing organization and their clients regard workers as independent contractors, turning them into mere factors of production, and their labor into a commodity or a product subject to market regulations \cite{Wood2019a}. They are paid a few cents of a dollar per task, mostly lack social protection traditionally tied with employment relations, and are subject to control and surveillance \cite{Tubaro2020,Gray2019, Irani2015}. Their assignments broadly comprise the interpretation and classification of data, and the work practices involved imply subjective social and technical choices that have significant yet hidden and vague ethical and political implications.  Workers interpreting and classifying data do not do so in a vacuum: their labor is embedded in large industrial structures and deeply intertwined with naturalized profit-oriented interests \cite{kazimzade_biased_2020}. How much room is left for horizontal discussion and agreement among data workers regarding how data is interpreted and sorted? Can the agency of workers prevail in such environments?

In this paper, we outline our on-going research that investigates the work of data annotators in Venezuela and Argentina. We discuss how the worldviews and interests of computer vision developers permeate datasets through commodified outsourced labor. Our research is based on 27 in-depth interviews with data annotators (on the outsourcing side) and computer vision practitioners (at the requesters' end), and the qualitative analysis of 210 annotation instruction documents formulated by requesters and carried out by outsourced workers.  Our preliminary findings show that, instead of seeking the “wisdom of crowds,” where a diverse and independent group cooperates to solve a problem, requesters of outsourced labor impose predefined forms of interpreting, classifying, and sorting data. Requesters exert their influence through narrow task instructions and specially tailored work interfaces. Managers and quality assurance analysts in BPOs, and algorithms in online platforms are in charge of overseing the process and making sure tasks are completed according to the needs of requesters. 

\section{Study Design and Method}

Through our research, we move away from approaches that aim at “identifying” and “mitigating” bias at the individual level and propose a broader view on this issue: We explore power asymmetries in data work and their effects on training datasets. Our research questions are: 

RQ1: How are ground truth labels for computer vision datasets instructed to annotators?

RQ2: What discourses are implicit in annotation instructions?

RQ3: What are the managerial practices related to those discourses?

As we describe in the following sub-sections, the exploration of these research questions comprises two qualitative methods: critical discourse analysis and in-depth interviewing.

\subsection{Discourse Analysis}
Task instructions provided by requesters to labelers play a crucial role in data annotation. They prescribe specific ways of interpreting data and the phenomena synthesized in that data. While task instructions significantly constrain the room for annotators’ biases, they constitute a fundamental tool to assure the imposition of computer vision companies’ values and interests on datasets. Annotators embed arbitrary preconceptions in training datasets by following said instructions. 

As part of this project, we plan to evaluate a corpus of 210 annotation instruction documents for computer vision data. We focus on instructions for annotation tasks carried out through three data annotation platforms in Venezuela and one business process outsourcing company in Argentina. Most of the documents comprise instructions related to facial recognition systems, autonomous vehicles, and scene understanding.
Our analysis focuses on the taxonomies at the core of pre-defined labels instructed by requesters. Through critical discourse analysis, we identify naturalized discourses encoded in them. The analysis of taken-for-granted assumptions embedded in discourses is fundamental because “Discourse (is) an interrelated set of texts, and the practices of their production, dissemination, and reception, that brings an object into being […] In other words, social reality is produced and made real through discourses, and social interactions cannot be fully understood without reference to the discourses that give them meaning.” \cite{phillips_discourse_2002} Critical discourse analysis explores the meanings that become dominant and those that are marginalized. Discourses encoded in annotation instructions will determine the realities that the dataset is able to reflect and the ones that have been erased from it.

We analyze discursive events using a three-dimensional framework \cite{bryman_social_2015}: The first dimension is that of the text, and corresponds with an examination of the actual content, structure, and meaning of each instruction document. The second dimension refers to the study of discursive practices.  Here, the focus is set on the form of discursive interaction used to communicate meaning and beliefs.  Finally, we examine the social practice dimension, i.e. the social contexts in which task instructions are composed and carried out.

\subsection{In-depth Interviews}
In addition to the discourse analysis, we have so far conducted 27 in-depth interviews with computer vision practitioners and data annotators working at crowdsourcing platforms and business process outsourcing companies. The goal of the interviews is revealing practices and perceptions as well as obtaining additional information about the contexts and relations that inform how annotation tasks come to be and how instructions are communicated.  The interviews include accounts of specific work situations involving the interpretation of data as well as the communication between annotators, managers, and requesters. Moreover, the interviews cover task descriptions, widespread routines and work practices as well as general views on the work and the data annotation and computer vision fields.

We use constructivist grounded theory principles \cite{charmaz_constructing_2006} to code and interpret the interview transcripts.  We follow phases of open, axial, and selective coding and let the categories emerge from the data. We apply a set of premises \cite{corbin_basics_2015} to make links between categories visible and make them explicit in our research documentation and in open discussions among the authors. We constantly compare the insights emerging from the interviews with the findings of our discourse analysis to revise our understanding or add additional evidence of observed phenomena.

\begin{table}
\begin{center}
\begin{tabular}{|c|c|c|c|}
\hline
Source & Type & Content & Number \\
\hline\hline
A      & BPO      & Various            & 6      \\
B      & Platform & Content Moderation & 32     \\
C      & Platform & Self-Driving Cars  & 152    \\
D      & Platform & Search Engines     & 20 \\
\hline
\end{tabular}
\end{center}
\caption{Sources of instructions}
\end{table}

\section{Preliminary Findings}

In concurrence with previous literature \cite{miceli_between_2020}, we have observed that workers interpreting, sorting, and labeling data do not do so in the vacuum of their personal judgment: their work and subjectivities are embedded in large industrial structures and subject to control. Power asymmetries in the production of datasets manifest in decisions related to what is considered data and how each data point is interpreted. We argue that possibilities for decision and interpretation in data work are constrained through top-down approaches. Artificial intelligence politics are inextricably connected to the power relations behind the collection and transformation of data and data work conditions, which allow preconceived hegemonic forms of knowledge to get encoded in computer vision systems via training datasets. 

In this case, the influence of the most powerful actors, i.e. computer vision companies and research institutions outsourcing the annotation assignments, permeates datasets through predefined forms of interpreting and sorting data. The annotation instructions reflect the particular worldviews of these actors. They require workers to categorize, label, enter data, and identify people, objects, and animals precisely, according to a set of preconceived ideas of how the world works and looks like. Thus, the taxonomies present in the instructions carry meanings that are self-evident to the requesters but not necessarily relevant to the annotators and their communities. For instance, most instructions to label images according to racial categories are based in US-centered conventions. For self-driving cars, annotation instructions refer to categories present in North American roads, such as signs and animals, that make little to no sense in Latin-American contexts.

Moreoever, annotation instructions reflect the commercial goals behind the computer vision systems that will be created on the base of annotated data. Predefined categories and labels mainly follow profit-oriented goals that prioritize classifications that have commercial application or are easier to operationalize in computational terms.  For instance, when asked what would be the procedure if they were unsure about annotation instructions, most of the workers at the Argentine BPO answered that they would immediately requesters' opinion because "their interpretation is usually the one that makes more sense as they know exactly what kind of system they're developing and how they plan to commercialize it."

Furthermore, the influence of powerful actors in data annotation is stabilized through narrow task instructions, specially tailored work interfaces, managers and quality assurance analysts in BPOs, and algorithms in labor platforms. Annotation instructions make reference to such control mechanisms. Instruction documents include warnings such as “low quality responses will be banned and not paid” or “accurate responses are required. Otherwise you will be banned.” These messages reinforce the annotation process’s hierarchical structure and compel workers to follow the norms as instructed. Data workers have little room to improve the labels or voice ethical concerns. They can only provide feedback on the aspects of the annotation process, that contribute to maintaining the status quo and power relations, such as pointing out a problem with the annotation interface or any other issues that could hinder the annotation process. Workers at the Argentine company are encouraged to think in terms of "what the client might want and what would bring more value to them". Here, any concerns expressed at the workers' end is filtered throughout hierarchical managerial structures and hardly ever reach requesters. Platform workers risk being banned and even expelled from the platform if they dare contradict task instructions. 

\section{Conclusion}

While the potentially harmful effects of algorithmic biases continue to be widely discussed, it is also essential to address how power imbalances and imposed classification principles in data creation contribute to the (re)production of inequalities by computer vision systems. The empowerment of workers and the decommodification of their labor away from market dependency, as well as the detailed documentation of outsourced processes of data creation, remain essential steps to allow spaces of reflection, deliberation, and audit that could potentially contribute to addressing some of the social questions surrounding computer vision.

{\small
\bibliographystyle{ieee_fullname}
\bibliography{egbib}
}

\end{document}